\def\year{2020}\relax
\documentclass[letterpaper]{article} 
\usepackage{aaai20}  
\usepackage{times}  
\usepackage{helvet} 
\usepackage{courier}  
\usepackage[hyphens]{url}  
\usepackage{graphicx} 
\usepackage{graphicx}  
\usepackage[linesnumbered,ruled,vlined]{algorithm2e}
\usepackage{multirow}
\usepackage{amsmath}
\usepackage{amssymb}
\usepackage{bm}
\usepackage{bbm, mathtools}
\usepackage{booktabs}
\title{Exploiting Motion Information from Unlabeled Videos for \\ Static Image Action Recognition}
\author{Yiyi Zhang,
Li Niu, \thanks{Corresponding author}
Ziqi Pan,
Meichao Luo,
Jianfu Zhang,
Dawei Cheng,
Liqing Zhang\footnotemark[1]\\
MoE Key Lab of Artificial Intelligence, Department of Computer Science and Engineering\\ Shanghai Jiao Tong University, Shanghai, China\\
\{yi95yi, ustcnewly, panziqi\_ai, luomcmxg, c.sis, dawei.cheng\}@sjtu.edu.cn, zhang-lq@cs.sjtu.edu.cn\\}

\begin{document}

\maketitle

\begin{abstract}
Static image action recognition, which aims to recognize action based on a single image, usually relies on expensive human labeling effort such as adequate labeled action images and large-scale labeled image dataset. In contrast, abundant unlabeled videos can be economically obtained. Therefore, several works have explored using unlabeled videos to facilitate image action recognition, which can be categorized into the following two groups: (a) enhance visual representations of action images with a designed proxy task on unlabeled videos, which falls into the scope of self-supervised learning; (b) generate auxiliary representations for action images with the generator learned from unlabeled videos. In this paper, we integrate the above two strategies in a unified framework, which consists of Visual Representation Enhancement (VRE) module and Motion Representation Augmentation (MRA) module. Specifically, the VRE module includes a proxy task which imposes pseudo motion label constraint and temporal coherence constraint on unlabeled videos, while the MRA module could predict the motion information of a static action image by exploiting unlabeled videos. We demonstrate the superiority of our framework based on four benchmark human action datasets with limited labeled data.
\end{abstract}

\section{Introduction}
Video action recognition has achieved competitive performance \cite{deepnet2,i3d}, but analyzing videos is computationally expensive. As an alternative approach, static image action recognition is becoming an active research topic \cite{human-body,context-scence,SC,reg,zhang2016action,im2flow}. 

However, image action recognition usually requires a large number of labeled action images. Although some image action datasets such as Pascal VOC \cite{VOC} and Stanford40 \cite{S40} are available, but they can only cover a limited number of action categories. In reality, not all categories have ample training samples, especially for those "long-tail" catgories \cite{slow-steady,SFA1}. Moreover, deep action image recognition approaches \cite{zhang2016action,im2flow} generally require CNN models pretrained on large-scaled labeled datasets (\emph{e.g.}, ImageNet~\cite{deng2009imagenet}) to obtain advanced visual representations of static images, while the collection of such datasets is very expensive and time-consuming. Instead, unsupervised visual representation learning is gaining momentum and attracting many researchers~\cite{zhang2017split,doersch2015unsupervised,noroozi2016unsupervised,gidaris2018unsupervised,slow-steady,odd-one-out,shuffle-learn,sort-seq}.

In this paper, we focus on static image action recognition with minimum supervision, \emph{i.e.}, only a few labeled action images and no pretrained CNN model. Despite the difficulty of this task, we humans are able to recognize the action in static images even after seeing only a few labeled action images, because our visual cortex can hypothesize motion signals from static image~\cite{bio2}. Inspired by neuropsychology, several works have attempted to learn motion cues from unlabeled videos to help image action recognition, because a large amount of unlabeled videos are freely available. These works can be divided into two research lines: (a) Enhance visual representations of action images with a proxy task based on unlabeled videos, which can capture supervisory signals in unlabeled videos such as temporal coherence \cite{SFA1,SFA2,slow-steady} and temporal order \cite{shuffle-learn,sort-seq,lstm}. This group of approaches are located in the realm of self-supervised learning, in which the main task is image action recognition and different proxy tasks are designed based on unlabeled videos; (b) Generate auxiliary representations for action images with the generator learned from unlabeled videos. For example, future visual representation and optical flow map are generated from action images in \cite{reg} and \cite{im2flow,dense-optical} respectively.

\begin{figure}[h]
\centering
\includegraphics[width=1.0\columnwidth]{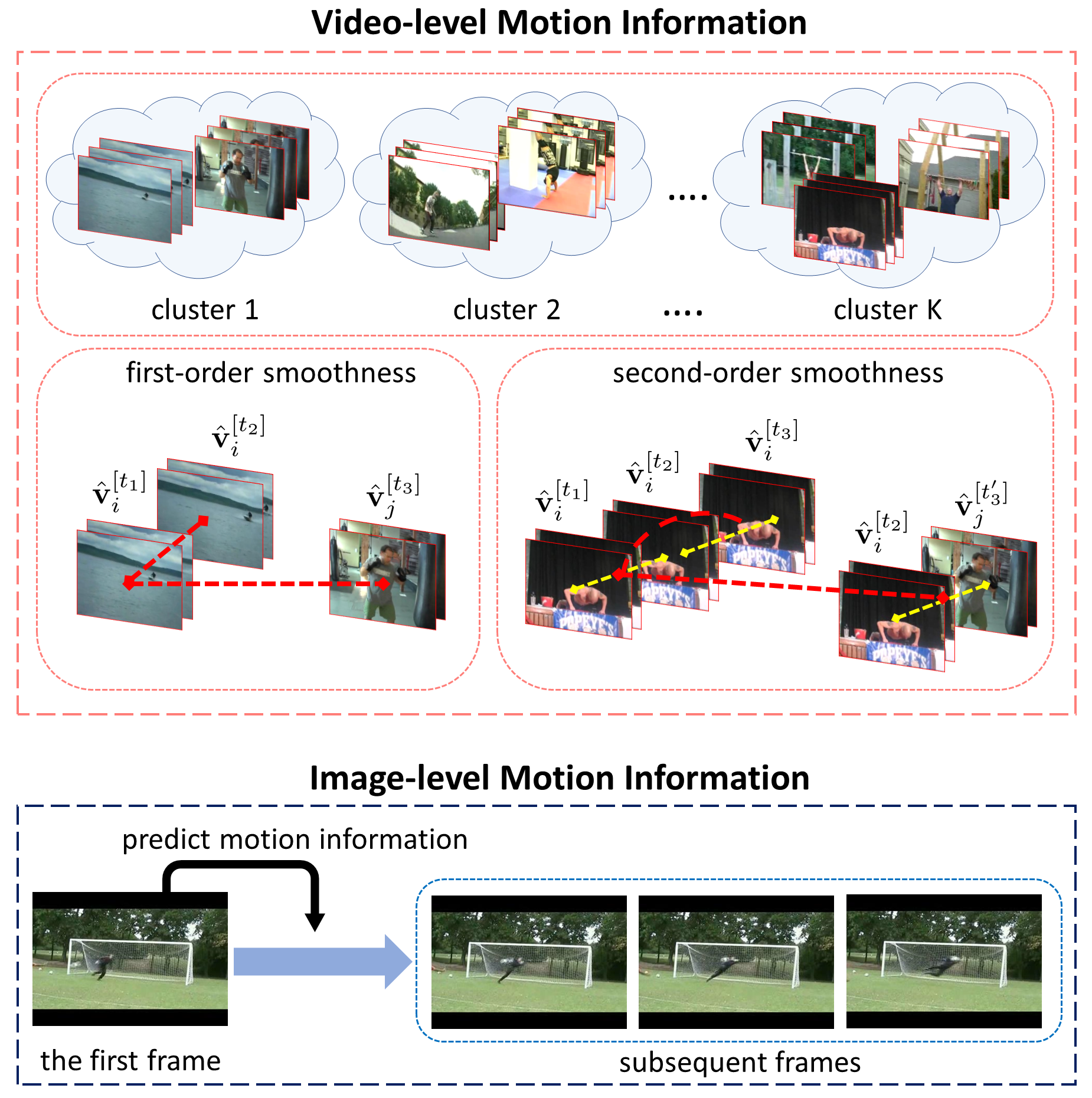}

\caption{Our framework exploits both video-level motion information and image-level motion information from unlabeled videos. On one hand, we exploit the discriminability and temporal coherence (\emph{i.e.}, first-order and second-order smoothness) of video-level motion information (see Section 3.1 for the definition of $\hat{\mathbf{v}}_i^{[t]}$). On the other hand, we exploit the image-level motion information by predicting the motion information of first frames in unlabeled videos. }
\label{fig1}
\end{figure}

One question is whether the above two research lines can be unified to fully exploit unlabeled videos. In this work, we explore this question by proposing a framework which can enhance the visual representations of action images and also generate auxiliary motion representations for action images. Our framework is composed of Visual Representation Enhancement (VRE) module and Motion Representation Augmentation (MRA) module:
(a) On one hand, our VRE module leverages video-level motion information from unlabeled videos to enhance visual representations of action images. Specifically, we design a proxy task based on the discriminability of video-level motion information, that is, unlabeled video clips could be classified into different clusters based on their pseudo motion labels (see Figure~\ref{fig1}). Inspired by \cite{slow-steady}, we also utilize the temporal coherence of video-level motion information, \emph{i.e.}, first-order and second-order smoothness between neighboring video clips (see Figure~\ref{fig1}), as supervisory signals; 
(b) On the other hand, our MRA module generates auxiliary motion representations for action images by exploiting the image-level motion information in unlabeled videos. Specifically, through predicting the motion information of the first frames in unlabeled videos (see Figure~\ref{fig1}), our MRA module is enabled to predict the motion information of static action images, which is used as auxiliary motion representations. Unlike \cite{im2flow,dense-optical} that generate low-level optical flow  from action images, we generate more robust high-level motion information from action images.
Finally, we concatenate the enhanced visual representations and the generated motion representations of action images for the final classification. 

Our contributions are threefold: 1) We propose a novel framework with VRE and MRA modules for static image action recognition, which unifies two research lines of exploiting unlabeled videos; 2) Our VRE and MRA module can take full advantage of video-level and image-level motion information from unlabeled videos; 3) Our framework outperforms the state-of-the-art image action recognition methods which exploit unlabeled videos, when minimum supervision is available. 

\section{Related Work}
In the section, we discuss the existing works which use unlabeled videos to help static image action recognition.


\subsection{Enhancing Visual Representation via Self-supervised Learning}
When using unlabeled videos to help image action recognition, many methods belong to self-supervised learning~\cite{slow-steady,odd-one-out,shuffle-learn,sort-seq}, which consists of a main task and a proxy task. Particularly, the main task is image action recognition and the proxy task is exploiting supervisory signals in unlabeled videos. Because the main task and the proxy task share the same visual representations, the supervisory signals of proxy task can help learn better visual representation in the main task. 

A diversity of proxy tasks have been designed to leverage unlabeled videos. 
Several methods \cite{SFA,SFA1,SFA2,slow-steady} investigated first-order or second-order smoothness, which means smooth transition between neighboring frames. Temporal order of video frames has also been studied in prior research \cite{Darwin,Wang,odd-one-out,shuffle-learn,sort-seq}.
Besides, there are some other types of proxy tasks using unlabeled videos. To name a few, the proxy task in \cite{pose-action} is to verify whether an optical flow map corresponds to a pair of starting and ending frames, while the proxy task in \cite{lstm} is using an encoder LSTM and a decoder LSTM to reconstruct video sequences.

\begin{figure*}[h]
\centering
\includegraphics[width=1.0\textwidth]{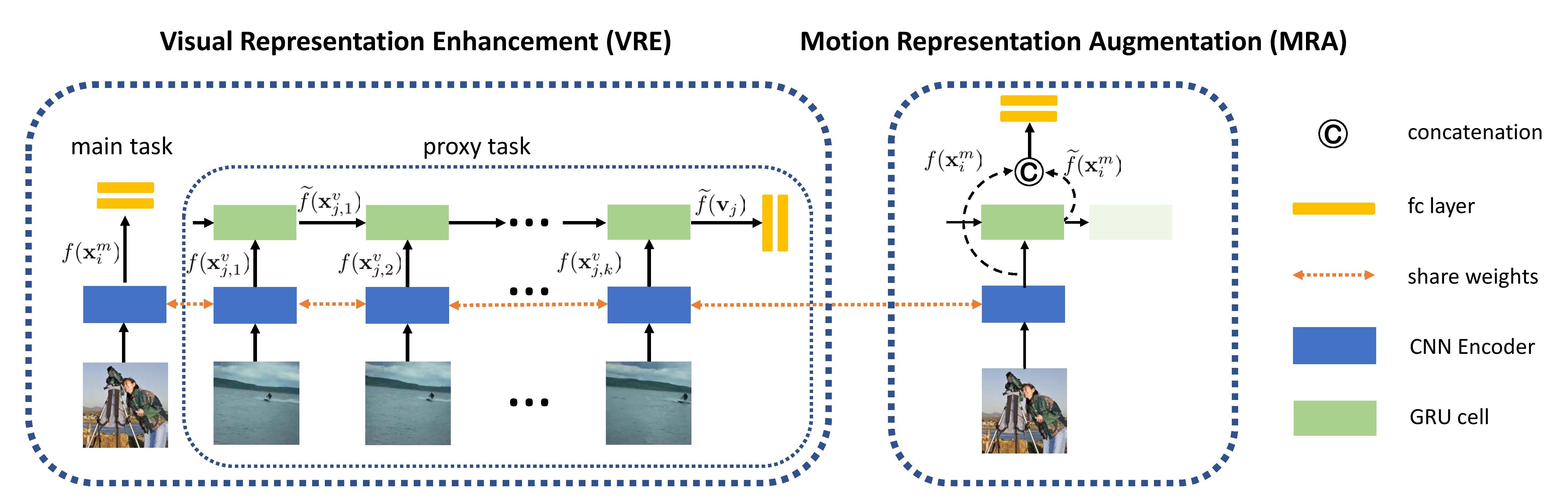}
\caption{Left: Our Visual Representation Enhancement (VRE) module is composed of a main task and a proxy task. The main task is image action recognition based on visual features $f(\mathbf{x}_i^m)$. The proxy task is video motion classification based on video-level motion features $\widetilde{f}(\mathbf{v}_j)$ with pseudo motion labels. All Encoders share the same model parameters and so do all GRU cells. 
Right: Our Motion Representation Augmentation (MRA) module generates image-level motion features $\widetilde{f}(\mathbf{x}_i^m)$ for static action images, which are concatenated with visual features $f(\mathbf{x}_i^m)$ for the final classification.}
\label{architecture}
\end{figure*}

Different from the above methods, video motion classification based on pseudo motion labels, i.e. the proxy task in our Visual Representation Enhancement (VRE) module, has not been explored in self-supervised learning previously. Although our proxy task also utilizes smoothness regularization similar to \cite{slow-steady}, our smoothness regularizer is imposed on video clips instead of frames to eliminate the disturbance of camera movement. Moreover, our framework can also generate auxiliary representations for action images to further improve the performance, which is out of the scope of self-supervised learning.

\subsection{Generating Auxiliary Information from Action Images} 
As another branch of using unlabeled videos for image action recognition, some methods aim to learn a generator based on unlabeled videos, and use the learned generator to produce auxiliary information from action images. For example, Carl \emph{etal.} \cite{reg} proposed to anticipate future visual representation of a static image. More works tended to predict motion information (\emph{e.g.}, optical flow, trajectory) of a static image \cite{dense-optical,im2flow,rf}.


The closest related work is \cite{dense-optical,im2flow}, in which motion information can be predicted from a static image. Their methods produce low-level optical flow map, while our Motion Representation Augmentation (MRA) module generates high-level motion representation that is more robust and discriminative. Moreover, our framework can also enhance visual representations of action images, which cannot be achieved by \cite{dense-optical,im2flow}.



\section{Our Method} \label{sec:our_approach}
For ease of presentation, in the remainder of this paper, we use a letter in boldface to denote a vector (\emph{e.g.}, $\mathbf{a}$) and a plain letter to denote a scalar (\emph{e.g.}, $a$). Besides, we use $\mathbf{a}^T$ to denote the transpose of vector $\mathbf{a}$. Assume we have a labeled action image dataset \(\mathcal{D}^m = \{(\mathbf{x}_i^m, \mathbf{y}_i)|_{i=1}^{N_m}\}\), in which $N_m$ is the number of images and $\mathbf{y}_i$ is the one-hot label vector of $\mathbf{x}_i^m$ with a single entry corresponding to its category label being $1$. 
Besides the labeled image dataset, we also have a set of unlabeled videos. For each unlabeled video, we sample the first $k$ key frames at the interval of $3$ frames to form a new video clip, in order to alleviate the influence of camera motion. Thus, we obtain a video clip dataset $\mathcal{D}^v = \{\mathbf{v}_i |_{i=1}^{N_v}\}$ with $\mathbf{v}_i = \{\mathbf{x}_{i,1}^v,\ldots, \mathbf{x}_{i,k}^v\}$. In the training stage, we tend to use both $\mathcal{D}^m$ and $\mathcal{D}^v$ to learn a robust action image classifier. As illustrated in Figure \ref{architecture}, our framework is composed of two modules: Visual Representation Enhancement (VRE) module and Motion Representation Augmentation (MRA) module.

\subsection{Visual Representation Enhancement (VRE) Module}  \label{sec:VRE}

In our VRE module, we enhance visual representations of action images by exploiting discriminative and smooth motion information of unlabeled videos. Following the terminology of self-supervised learning, our main task is image action recognition and our proxy task is video motion classification with pseudo motion labels.

\textbf{Image Action Recognition as Main task: }Our main task is static image action recognition and we use CNN model (\emph{e.g.}, AlexNet) with random initialization as our backbone network. By taking AlexNet~\cite{AlexNet} as an example, we denote AlexNet without the last two fully connected (fc) layers as a visual feature extractor $f(\cdot)$ with model parameters $\bm{\theta}_n$, which is also referred to as CNN Encoder in the remainder of this paper. Thus, the visual feature of action image $\mathbf{x}_i^m$ can be represented by $f(\mathbf{x}_i^m)$. Besides, we denote the last two fc layers of AlexNet as a visual classifier $p_c(\cdot)$ with model parameters $\bm{\theta}_c$. Then, the cross-entropy classification loss of main task can be written as
\begin{equation}
    L_{main} = -\sum_{i=1}^{N_m}\mathbf{y}_i^T\log p_c(f(\mathbf{x}_i^m)).
\label{eq:cross-entropy}
\end{equation}

\textbf{Video Motion Classification as Proxy Task: }Our proxy task is video motion classification with pseudo motion labels, where the discriminability of video-level motion information of unlabeled videos can be exploited. 
We cluster unlabeled videos using K-means based on their hand-crafted features ($K=16$ in our experiments) and treat the cluster label of each video clip as its pseudo motion label.  The details of hand-crafted features are left to Supplementary.
To perform video motion classification based on the obtained pseudo motion labels,  we employ Recurrent Neural Network (RNN) architecture with multiple cells sharing the same model parameters, as shown in Figure \ref{architecture}. Specifically, we adopt Gated Recurrent Unit (GRU) cell \cite{GRU} with model parameters $\bm{\theta}_g$ in RNN architecture. Given a video clip $\mathbf{v}_j$, the $t$-th GRU cell takes the output  from the $(t-1)$-th GRU cell which is denoted as $\widetilde{f}(\mathbf{x}_{j,t-1}^v)$ and the visual feature $f(\mathbf{x}_{j,t}^v)$ of the $t$-th frame as input. 
Video motion classification is based on the output $\widetilde{f}(\mathbf{x}_{j,k}^v)$ from the last GRU cell, which is also denoted as $\widetilde{f}(\mathbf{v}_j)$.  $\widetilde{f}(\mathbf{v}_j)$ is supposed to represent the video-level motion feature because of the supervision by pseudo motion labels.

By using $\hat{\mathbf{y}}_i$ to denote the one-hot  pseudo motion label vector of $\mathbf{v}_i$, the cross-entropy classification loss can be written as
\begin{eqnarray}
    L_{motion} = -\sum_{i=1}^{N_v}\hat{\mathbf{y}}_i^T\log p_m(\widetilde{f}(\mathbf{v}_i)),
    \label{eq:main}
\end{eqnarray}
in which $p_m(\cdot)$ is the motion classifier consisting of two fully connected (fc) layers with model parameters $\bm{\theta}_m$.

The proxy task shares the same CNN Encoder with the main task, so the supervision information from the proxy task can contribute to better visual representation in the main task. 

\textbf{Video-level Smoothness Regularization: }Inspired by \cite{slow-steady}, we additionally exploit the temporal coherence of unlabeled videos to further enhance visual representations. Unlike \cite{slow-steady}, we impose smoothness constraints on video-level motion features in lieu of image-level visual features, to mitigate the noise induced by camera motion.

First, we segment some shorter video clips based on $\mathcal{D}^v=\{\mathbf{v}_i|_{i=1}^{N_v}\}$. Specifically, we use \(\hat{\mathbf{v}}_i^{[t]} = \{\mathbf{x}_{i,t}^v, \cdots, \mathbf{x}_{i,t+\Delta t-1}^v\}\) to denote a shorter video clip with the starting frame $t$ and  duration \(\Delta t\), which is segmented from \(\mathbf{v}_i\).  
We collect the set of temporal neighbor pairs $\mathcal{S}_1 = \{(\hat{\mathbf{v}}_i^{[t_1]}, \hat{\mathbf{v}}_i^{[t_2]}) \mid_{i=1}^{N^v} |t_2-t_1|\leq \lceil\Delta t/2\rceil\}$,
which means that in a temporal neighbor pair, two video clips should be from the same $\mathbf{v}_i$ and the overlap between them should be at least half of their duration $\Delta t$. We impose first-order smoothness regularization on $\mathcal{S}_1$, which can be formulated as contrastive loss \cite{contrast} as follows,
\begin{equation}
\begin{split}
    L_{smooth1} &=\sum_{(\hat{\mathbf{v}}_i^{[t_1]}, \hat{\mathbf{v}}_i^{[t_2]})\in \mathcal{S}_1}\!\!\!\!\! d\left(\widetilde{f}(\hat{\mathbf{v}}_i^{[t_1]}), \widetilde{f}(\hat{\mathbf{v}}_i^{[t_2]})\right)\\
   & \!\!\!\!\!\!\!\!\!\!\!\!\!\!\!\!\!\!\!\!\!\!\!\!\!\!\!\!\!\!+\sum_{(\hat{\mathbf{v}}_i^{[t_1]}, \hat{\mathbf{v}}_j^{[t_2]})\notin \mathcal{S}_1}\!\!\!\!\!\max\left(\delta-d\left(\widetilde{f}(\hat{\mathbf{v}}_i^{[t_1]}), \widetilde{f}(\hat{\mathbf{v}}_j^{[t_2]})\right),0\right),
\end{split} \nonumber
\end{equation}
in which $d(\cdot, \cdot)$ measures the Euclidean distance and the margin $\delta$ is set to $1$ in our experiments. $\widetilde{f}(\mathbf{v})$ is the video-level motion feature of $\mathbf{v}$, which is extracted by the RNN architecture consisting of $\Delta t$ GRU cells. 

In analogy to temporal neighbor pairs, we collect the set of temporal neighbor tuples $\mathcal{S}_2 = \{(\hat{\mathbf{v}}_i^{[t_1]}, \hat{\mathbf{v}}_i^{[t_2]}, \hat{\mathbf{v}}_i^{[t_3]}) \mid_{i=1}^{N^v} |t_2-t_1|\leq \lceil\Delta t/2\rceil, |t_3-t_2|\leq \lceil\Delta t/2\rceil\}$, which means that in a temporal neighbor tuple, three video clips should be from the same $\mathbf{v}_i$, and they should be overlapped by at least half of their duration $\Delta t$. Besides, for each tuple in $\mathcal{S}_2$, we replace the third element with a video clip $\hat{\mathbf{v}}_j^{[t_3]}$ from another $\mathbf{v}_j$, leading to another set of tuples
$\hat{\mathcal{S}}_2 = \{(\hat{\mathbf{v}}_i^{[t_1]}, \hat{\mathbf{v}}_i^{[t_2]}, \hat{\mathbf{v}}_j^{[t_3]}) \mid_{i,j=1}^{N^v}  i \neq j, |t_2-t_1|\leq \lceil\Delta t/2\rceil\}$.

By denoting $\hat{d}(t_1,t_2)=\widetilde{f}(\hat{\mathbf{v}}_i^{[t_1]})-\widetilde{f}( \hat{\mathbf{v}}_i^{[t_2]})$, we impose second-order smoothness regularization on $\mathcal{S}_2$ and $\hat{\mathcal{S}}_2$ by using contrastive loss: 
\begin{align}
\begin{split}
   L_{smooth2} & = \!\!\!\!\!\!\!\!\!\sum_{(\hat{\mathbf{v}}_i^{[t_1]}, \hat{\mathbf{v}}_i^{[t_2]}, \hat{\mathbf{v}}_i^{[t_3]}) \in \mathcal{S}_2} \!\!\!\!\!\!\!\!\!\!\!\! d\left(\hat{d}(t_1,t_2), \hat{d}(t_2,t_3)\right) \\
    &\!\!\!\!\!\!\!\!\!\!\!\!\!\!\!\!\!\!\!\!\!\!\!\!\!\!\!\!\!+ \!\!\!\!\!\!\!\!\!\!\!\!\!\!\sum_{\quad\quad(\hat{\mathbf{v}}_i^{[t_1]}, \hat{\mathbf{v}}_i^{[t_2]}, \hat{\mathbf{v}}_j^{[t_3]}) \in \hat{\mathcal{S}}_2} \!\!\!\!\!\!\!\!\!\!\!\!\!\max\!\left(\delta\!-\!d\left(\hat{d}(t_1,t_2),  \hat{d}(t_2,t_3)\right),0\right),
\nonumber
\end{split}
\end{align}
which has similar explanation to $L_{smooth1}$. After incorporating smoothness regularization, the loss of the proxy task in our VRE module can be written as
\begin{equation}
   L_{proxy}=L_{motion} + L_{smooth1} +  L_{smooth2}.
  \label{eq:proxy}
\end{equation}
Then, we can arrive at the total loss of our VRE module 
\begin{equation}
  L_{VRE} = L_{main} + \lambda L_{proxy},
  \label{eq:all}
\end{equation}
where $\lambda$ is a trade-off parameter and set to $0.1$ in our experiments.

\subsection{Motion Representation Augmentation (MRA) Module} \label{sec:MRA}
After the RNN architecture is trained in our VRE module, we assume that the output from the last GRU cell represents the video-level motion information of the input video clip due to the supervision of pseudo motion labels.
In our MRA module, we only use one GRU cell learned in our VRE module to generate motion representation for a static action image. Note that in the proxy task of our VRE module, the first GRU cell uses a constant vector (\emph{i.e.}, all-zero vector) to supersede the output from previous GRU cell, which means that the first frame in a video does not have prior motion information from previous frames. Actually, a static action image $\mathbf{x}_i^m$ can be treated as the first frame of an imaginary video clip with its prior motion information unknown. Therefore, we feed the same constant all-zero vector and $f(\mathbf{x}_i^m)$ into a GRU cell, as illustrated in Figure \ref{architecture}. We assume that the output $\widetilde{f}(\mathbf{x}_i^m)$ from this GRU cell could represent the high-level motion information of $\mathbf{x}_i^m$, which is visualized and verified in our experiments.

\begin{algorithm}[t]
 \SetKwInOut{Input}{Input}
\Input{Labeled images $\mathcal{D}^m=\{(\mathbf{x}_i^m, \mathbf{y}_i)|_{i=1}^{N_m}\}$ and unlabeled video clips  $\mathcal{D}^v=\{\mathbf{v}_i |_{i=1}^{N_v}\}$.} 
Acquire pseudo motion labels for $\mathcal{D}^v=\{\mathbf{v}_i |_{i=1}^{N_v}\}$.\\
Initialize $\{\bm{\theta}_c, \bm{\theta}_n, \bm{\theta}_g, \bm{\theta}_m, \bm{\theta}_a\}$ randomly.\\
\tcc{Visual Representation Enhancement Module}
Update $\{\bm{\theta}_n, \bm{\theta}_c\}$ based on $L_{main}$ in (\ref{eq:cross-entropy}).\\
Update $\{\bm{\theta}_n, \bm{\theta}_c, \bm{\theta}_g, \bm{\theta}_m\}$ based on $L_{VRE}$ in (\ref{eq:all}).\\
Update $\{\bm{\theta}_n, \bm{\theta}_c\}$ based on $L_{main}$ in (\ref{eq:cross-entropy}).\\
\tcc{Motion Representation Augmentation Module}
Update $\{\bm{\theta}_g, \bm{\theta}_m\}$ based on $L_{proxy}$ in (\ref{eq:proxy}).\\
Update $\bm{\theta}_a$ based on $L_{MRA}$ in (\ref{eq:fusion}).\\
\textbf{return} $\{\bm{\theta}_n, \bm{\theta}_g, \bm{\theta}_a\}$
%
%
 \caption{Training procedure of our framework.}
 \label{algo1}
\end{algorithm}

Although CNN Encoder is also supervised by unlabeled videos in the proxy task, it focuses on the main task of image action recognition. So we claim that CNN Encoder produces visual representations which are enhanced by the proxy task. In contrast, GRU cell is only supervised by unlabeled videos and focuses on learning transferable motion representation. So the generated motion representations of action images are complementary with their visual representations to some extent. Therefore, the combination of both is likely to boost the performance in action recognition.

For each action image, we concatenate its visual representation  $f(\mathbf{x}_i^m)$ and motion representation $\widetilde{f}(\mathbf{x}_i^m)$ as $[\widetilde{f}(\mathbf{x}_i^m); f(\mathbf{x}_{i}^m)]$, followed by two fully connected (fc) layers with model parameters $\bm{\theta}_a$ which play the role as the final classifier $p_a(\cdot)$. The cross-entropy classification loss can be written as
\begin{equation}
    L_{MRA} =- \sum_{i=1}^{N_m}\mathbf{y}_i^T\log p_a([\widetilde{f}(\mathbf{x}_i^m); f(\mathbf{x}_{i}^m)]),
\label{eq:fusion}
\end{equation}
which resembles $L_{main}$ in (\ref{eq:cross-entropy}) except that (\ref{eq:fusion}) uses $[\widetilde{f}(\mathbf{x}_i^m); f(\mathbf{x}_{i}^m)]$ instead of $f(\mathbf{x}_{i}^m)$.


\paragraph{\textbf{Training Procedure: }}
After randomly initializing all model parameters, our whole training procedure has five steps: (1) Initialize CNN model (\emph{i.e.}, $\{\bm{\theta}_n, \bm{\theta}_c\}$) based on the main task in our VRE module, in which only labeled action images are used; (2) Train CNN model (\emph{i.e.}, $\{\bm{\theta}_n, \bm{\theta}_c\}$), GRU cell (\emph{i.e.}, $\bm{\theta}_g$), and motion classifier (\emph{i.e.}, $\bm{\theta}_m$) based on the main task and the proxy task in our VRE module, in which both labeled action images and video clips with pseudo motion labels are used;  (3) Fine-tune CNN model (\emph{i.e.}, $\{\bm{\theta}_n, \bm{\theta}_c\}$) based on the main task in our VRE module, which is similar to the first step. Up to now, we hope to obtain satisfactory visual representations for action images; (4) Fine-tune GRU cell (\emph{i.e.}, $\bm{\theta}_g$) and motion classifier (\emph{i.e.}, $\bm{\theta}_m$) based on the proxy task in our VRE module, in which only video clips with pseudo motion labels are used. Up to now, we expect to obtain satisfactory motion representations for action images; (5) Train the classifier (\emph{i.e.}, $\bm{\theta}_a$) based on the concatenated visual and motion representations, in which only labeled action images are used. In the testing stage, given a test image, we pass it through the CNN Encoder and GRU cell to obtain its visual representation and motion representation, which are concatenated for the final prediction.

\section{Experiments}
We evaluate our framework through extensive experiments on two action image datasets and two unlabeled video datasets.

\subsection{Experimental Setup}
We use two action image datasets (\emph{i.e.}, PASCAL VOC and Stanford40) and two video datasets (\emph{i.e.}, UCF101 and HMDB), which can form four pairs of video\(\rightarrow\)image datasets corresponding to four experimental settings (\emph{i.e.}, UCF101\(\rightarrow\)VOC, UCF101\(\rightarrow\)Stanford40, HMDB\(\rightarrow\)VOC, and HMDB\(\rightarrow\)Stanford40). 

\textbf{Action image datasets:} PASCAL VOC~\cite{VOC} is an action image dataset with $10$ categories and Stanford40 \cite{S40} contains $40$ daily human actions. Since this work focuses on minimum supervision, we follow \cite{slow-steady} to use only a few labeled training images per category. On VOC dataset, we use $10$ labeled training images per category and $2000$ images for testing. On the Stanford40 dataset, we use $30$ labeled training images per category and standard $5532$ test images specified in \cite{S40}. Besides using minimum supervision which is the focus of this paper, we also investigate using more training samples per category in Section 4.6.

\textbf{Video datasets:} HMDB \cite{HMDB} contains 6849 short video clips distributed in 51 actions and UCF101 \cite{UCF101} contains 13320 videos from 101 action categories. Note that we do not use the provided action labels of videos. Following \cite{slow-steady}, we randomly sample 1000 video clips from both datasets, \emph{i.e.}, $N_v=1000$. We extract the first $k=12$ (\emph{resp.}, $k=7$) key frames from each video at the interval of $3$ frames on the UCF101 (\emph{resp.}, HMDB) dataset considering the difference of video lengths between these two datasets. For video motion classification, we split each $k$-frame video clip into three $(k-2)$-frame video clips to augment training data.
For smoothness regularization, $\Delta t$ is set as $10$ (\emph{resp.}, $5$) on the UCF101 (\emph{resp.}, HMDB) dataset.

\textbf{Implementation details:} We use AlexNet as backbone by default. The  visual representation from CNN Encoder is $4096$-dim and the motion representation from GRU cell is $512$-dim. The motion classifier $f_m(\cdot)$ in our VRE module has two fully connected layers with intermediate $512$-dim output. The classifier $f_a(\cdot)$ based on concatenated $4608$-dim features in MRA module also has two fully connected layers with intermediate $4608$-dim output. During training, we use Adam optimizer, in which the learning rate starts with $0.0001$ and decays by $0.1$ every $1800$ iterations. 

\begin{table*}[h]
\caption{Accuracies (\%) of different methods in four experimental settings. The best results are denoted in boldface.}
\centering
\small
\begin{tabular}{c|cccc} 
\toprule[1pt]
 Datasets & UCF101\(\rightarrow\)VOC & UCF101\(\rightarrow\)S40 & HMDB\(\rightarrow\)VOC & HMDB\(\rightarrow\)S40 \\ \hline
UNREG  & 16.22 & 15.42  & 16.22 & 15.42 \\
O3N \cite{odd-one-out} &16.28 & 16.36 & 16.28 & 16.79 \\
\citeauthor{sort-seq} \shortcite{sort-seq}  &18.00 & 17.19 & 16.89 & 16.52\\
OPN \cite{shuffle-learn}& 18.44 & 16.83 & 18.39 & 16.63 \\
\citeauthor{lstm} \shortcite{lstm} & 16.42 & 15.99 & 18.39 & 16.42\\
\citeauthor{pose-action} \shortcite{pose-action} &16.55 &16.06 & 16.55&16.89\\
SSFA \cite{slow-steady}  & 15.11  & 16.27  & 17.06  & 17.66\\
\hline
\citeauthor{reg} \shortcite{reg} &19.28 &17.42 &17.94 &16.87 \\
\citeauthor{im2flow} (2s) \shortcite{im2flow} &17.94 &15.29 &16.84 &14.67 \\
\citeauthor{im2flow} \shortcite{im2flow} &18.39 &16.87 &17.94 &15.89 \\
\citeauthor{rf} \shortcite{rf} &18.39 &15.93 &19.17 &17.08 \\
\citeauthor{dense-optical} \shortcite{dense-optical} &19.11 &17.44 &17.56 &15.69 \\
\hline
\textbf{Ours} &\textbf{21.06} &\textbf{18.76} & \textbf{22.94} &\textbf{19.11}\\
\bottomrule[1pt]
\end{tabular}
\label{table:2}
\end{table*}

\begin{table}[t]
\centering
\small
\caption{Accuracies (\%) of our special cases in two experimental settings. }
\begin{tabular}{c |c |c  } 
\toprule[1pt]
 Datasets  & UCF101\(\rightarrow\)VOC  & HMDB\(\rightarrow\)VOC  \\\hline
UNREG  & 16.22  & 16.22  \\
UNREG+motion &19.39  & 19.72  \\
UNREG+smooth1 &18.89    & 18.61 \\
UNREG+smooth2  &18.72    & 18.11 \\
Ours (w/o MRA) & 20.06  & 20.44  \\
only MR &13.72  &18.17  \\
Ours &21.06   &22.94  \\\hline
SUP-FT  & 28.72  & 28.72 \\
\bottomrule[1pt]
\end{tabular}
\label{table:3}
\end{table}

\subsection{Baselines}
We compare with two groups of state-of-the-art baselines: 1) The first group of baselines design different proxy tasks based on unlabeled videos to enhance visual representations, which falls into the scope of self-supervised learning. Specifically, we compare with O3N \cite{odd-one-out}, OPN \cite{shuffle-learn}, Misra \emph{etal.} \cite{sort-seq}, SSFA \cite{slow-steady}, Srivastava \emph{etal.} \cite{lstm}, and Purushwalkam \emph{etal.} \cite{pose-action}; 2) The second group of baselines generate auxiliary representations for action images, with the generator learned based on unlabeled videos. We compare with Carl \emph{etal.} \cite{reg}, which can generate future visual representation given a static action image. For each image, we concatenate its current visual representation and future visual representation, followed by two fully connected layers for classification. 
Moreover, we compare with Jacob \emph{etal.} \cite{dense-optical}, Pintea \emph{etal.} \cite{rf}, and Gao \emph{etal.} \cite{im2flow}, which can generate optical flow map given a static action image. We extract features from their generated optical flow maps in a similar way to our hand-crafted video features, and then concatenate them with visual representations for the final classification. For Gao \emph{etal.} \cite{im2flow}, we additionally adopt two-stream network \cite{two-stream} by combining spatial stream CNN and temporal stream CNN for prediction score fusion, which is referred to as Gao \emph{etal.} (2s) in Table \ref{table:2}.

Additionally, we include the baseline UNREG, which stands for UNREGularized network. Specifically, we train a CNN model from scratch by using labeled training images based on $L_{main}$ in (\ref{eq:main}), without the help of unlabeled videos.

\subsection{Experimental Results} \label{sec:cmp_baseline}
The results of different methods are reported in Table~\ref{table:2}, from which we observe that
the first group of baselines 
are generally better than UNREG, which proves that it is helpful to use various proxy tasks based on unlabeled videos to enhance visual representations. The second group of baselines also generally outperform UNREG,
which indicates the advantage of future visual representation or auxiliary motion information. Finally, our framework outperforms two groups of baselines and achieves the best results in all four experimental settings, which corroborates the effectiveness of enhancing visual representations by video motion classification and generating auxiliary motion representations.


\subsection{Ablation Studies} \label{sec:ablation}
To validate each component of our framework, we perform extensive ablation studies on our special cases in this section. Firstly, we focus on our VRE module and verify the effectiveness of each loss term (\emph{i.e.}, $L_{motion}$, $L_{smooth1}$, $L_{smooth2}$). In particular, we perform the first three steps in the training procedure and replace $L_{proxy}$ in the second step with only $L_{motion}$ (\emph{resp.}, $L_{smooth1}$, $L_{smooth2}$), which is referred to as UNREG+motion (\emph{resp.}, UNREG+smooth1, UNREG+smooth2) in Table \ref{table:3}. Similarly, we perform the first three steps in the training procedure and use full $L_{proxy}$ in the second step, which is referred to as Ours (w/o MRA) in Table \ref{table:3} because this special case does not use the MRA module.
Secondly, we focus on the MRA module and test the performance only using auxiliary Motion Representation (MR), which is referred to as ``only MR" in Table \ref{table:3}. In this special case, we need to learn a classifier based on auxiliary motion representations.  

We also include the baseline SUP-FT, which stands for SUPervised-pretrained and Fine-Tune. Specifically, we fine-tune the CNN model pretrained on the ImageNet dataset using the labeled action training images based on $L_{main}$ in (\ref{eq:main}), without the help of unlabeled videos. 

By taking UCF101\(\rightarrow\)VOC and HMDB\(\rightarrow\)VOC as examples, the results of all special cases are summarized in Table \ref{table:3}. By comparing UNREG+motion with UNREG, we observe that the proxy task of video motion classification with pseudo motion labels plays a crucial role in our VRE module, which achieves 2.86\% improvement over UNREG on average. Besides, first-order (\emph{resp.}, second-order) smoothness regularization are also helpful when comparing UNREG+smooth1 (\emph{resp.}, UNREG+smooth2) with UNREG. The performance only using auxiliary motion representations is better than random guess, and even better than UNREG in HMDB\(\rightarrow\)VOC setting, suggesting that the GRU cell could generate meaningful motion representations.
It can also been seen that our full-fledged method outperforms Ours (w/o MRA), which demonstrates that it is beneficial to augment visual representations with motion representations.

Based on Table \ref{table:3}, there is a performance gap between our method and SUP-FT. However, SUP-FT is pretrained on ImageNet with millions of  labeled images which requires sheer expense of human annotation, while our method only utilizes minimum supervision.



\begin{figure}[t]
\centering
\includegraphics[width=0.8\columnwidth]{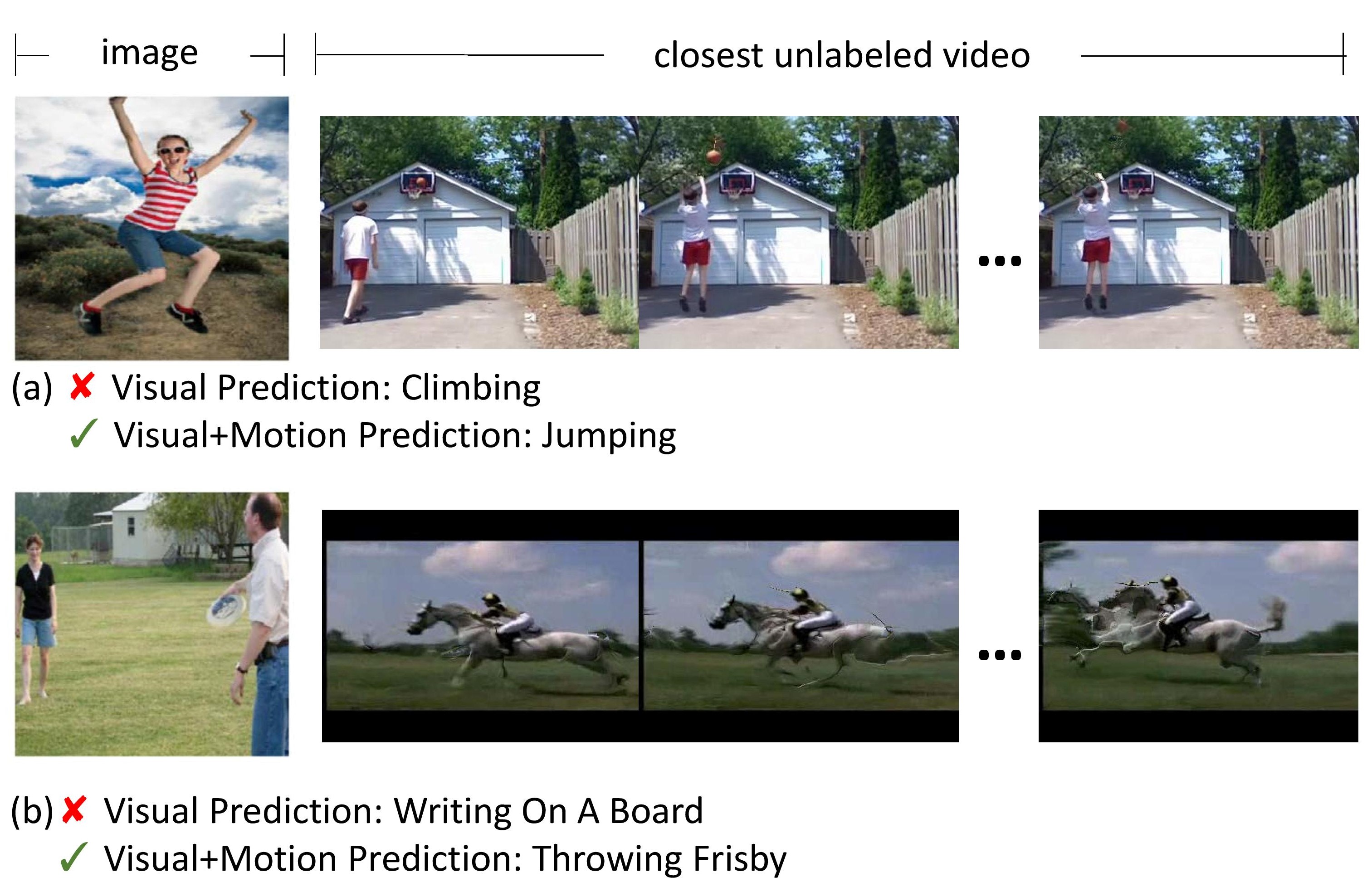}
\caption{Images are misclassified based on visual representation, but are correctly classified based on both visual representation and motion representation.}
\label{viz1}
\end{figure}

\subsection{Qualitative Analyses of Motion Representation} \label{sec:qualitative_analyses}
To visualize auxiliary motion representation (MR) of action images, for each action image $\mathbf{x}_i^m$, we show the video clip whose first frame has the closest MR to this action image. Specifically, after comparing \(\widetilde{f}(\mathbf{x}_i^m)\) of image $\mathbf{x}_i^m$ with  \(\widetilde{f}(\mathbf{x}^v_{j,1})\) of all unlabeled video clips \(\mathbf{v}_{j}\) in $\mathcal{D}^v$, we retrieve the video clip \(\mathbf{v}_{j^*}\) with minimum \(||\widetilde{f}(\mathbf{x}_i^m)-\widetilde{f}(\mathbf{x}^v_{{j^*},1})||_2\). The retrieved video clips are assumed to visually represent the motion information encoded in MR.

In Figure~\ref{viz1}, we observe that auxiliary MR can rectify wrong prediction. For example, the image in (b) is misclassified as ``writing on a board" based on visual representation, probably because the frisby visually resembles a white board. However, the motion information in MR, \emph{i.e.}, similar hand movement in ``horse riding", helps classify this image as ``throwing frisby" correctly.
The experimental results show that MR can capture the motion information of static action images and compensate the drawback of visual representation. 
More results are provided in Supplementary.

\subsection{Impact of Different Numbers of Labeled Action Images}
\begin{figure} [t]

    \centering
    \resizebox{.7\columnwidth}!{\includegraphics[scale=.42]{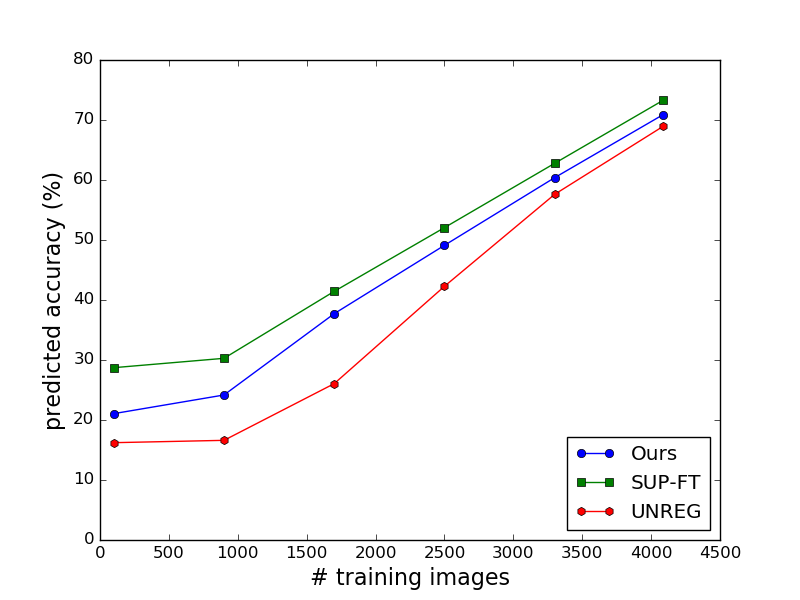}}
    \caption{Accuracies (\%) of UNREG, SUP-FT, and our framework in UCF101\(\rightarrow\)VOC setting with different numbers of labeled training images.}
    \label{table:5}
\end{figure}

We have demonstrated that exploiting motion information from unlabeled videos is beneficial when the number of labeled training images is limited. It is interesting to investigate whether exploiting motion information from unlabeled videos is still useful when the number of labeled training images increases. By taking UCF101 \(\rightarrow\) VOC as an example, we vary the number of labeled training images in [100, 900, 1700, 2500, 3300, 4087], in which $4087$ is the maximum number of available training images.  The performance of UNREG, SUP-FT, and our framework is reported in Figure \ref{table:5}, from which it can be seen that our framework is consistently effective when using different numbers of labeled training images. When the number of labeled training images is very large (\emph{e.g.}, 4087), the performance of UNREG is approaching SUP-FT, which means that labeled action images already have sufficient supervision information. In this case, the performance gain brought by our framework is also becoming marginal.

\subsection{Impact of Different Network Structures}
To demonstrate the generalizability of our framework, we replace the AlexNet backbone with VGG16~\cite{simonyan2014very}. Experimental results are left to Supplementary due to space limitation.

\section{Conclusion}
In this paper, we proposed a novel framework, which unifies two research lines of exploiting unlabeled videos to help static image action recognition: Visual Representation Enhancement (VRE) module and Motion Representation Augmentation (MRA) module. Comprehensive experimental results have revealed the superiority of our framework.

\section{Acknowledgement}
This work is supported by the National Key R\&D Program of China (2018AAA0100704) and is partially sponsored by Shanghai Sailing Program (BI0300271) and Startup Fund for Youngman Research at SJTU (WF220403041).

\newcommand{\beginsupplement}{%
        \setcounter{table}{0}
        \renewcommand{\thetable}{S\arabic{table}}%
        \setcounter{figure}{0}
        \renewcommand{\thefigure}{S\arabic{figure}}%
        \setcounter{section}{0}
        
        }

\section{Supplementary}
\beginsupplement
\section{Hand-crafted Video Representation} 
In our VRE module, we need to obtain the pseudo labels of video clips $\mathcal{D}^v=\{\mathbf{v}_i|_{i=1}^{N_v}\}$ to perform video motion classification, which is achieved by clustering based on their hand-crafted representations. 
For each key frame $\mathbf{x}_{i,t}^v$ in each video clip $\mathbf{v}_i$, we compute the optical flow map between $\mathbf{x}_{i,t}^v$ and $\mathbf{x}_{i,t+1}^v$ using Coarse2Fine \cite{course2fine}, which is an efficient unsupervised method. Each optical flow map contains an entry \((dx, dy)\) for the image pixel $(x, y)$, which indicates the movement of this pixel. Then, we calculate an OTSU threshold \cite{otsu} based on the magnitude $\sqrt{dx^2+dy^2}$ of all entries \((dx, dy)\) in all optical flow maps and the entries with the magnitude below the threshold are discarded, resulting in the filtered optical flow maps. Then, $128$ clusters are learned with K-means \cite{kmeans} based on all the remaining entries in all optical flow maps. By using the learned clusters as the codebook, we use Vector of Locally Aggregated Descriptors (VLAD) \cite{vlad} to encode each filtered optical flow map into an image-level VLAD embedding with the dimensionality $128\times 2=256$. 
Finally, for each video clip $\mathbf{v}_i$, we concatenate the image-level VLAD embeddings of all its key frames $\mathbf{x}_{i,t}^v$, leading to a video-level VLAD embedding of $\mathbf{v}_i$.

\begin{figure}[b]
\centering
\includegraphics[width=0.9\columnwidth]{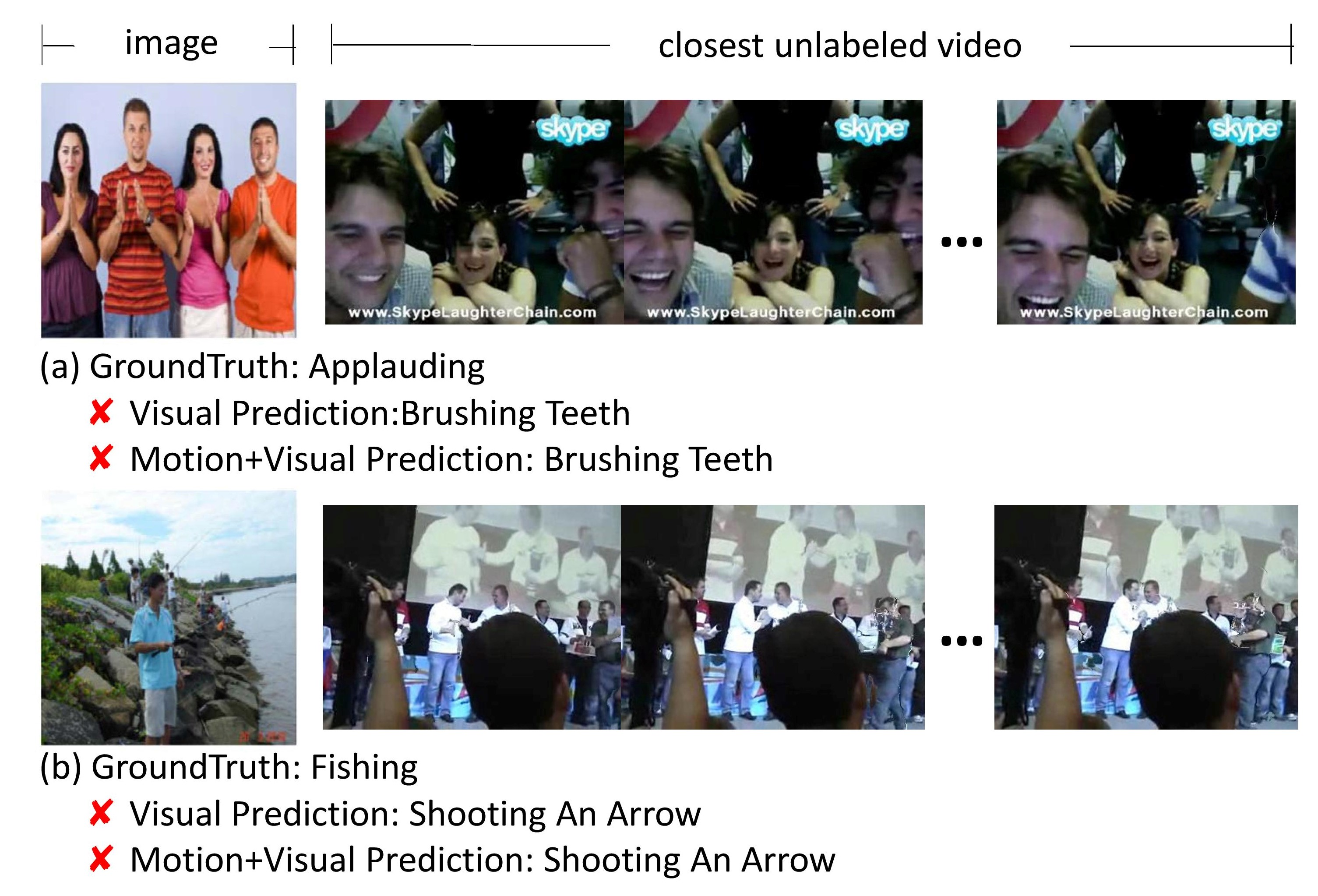}
\caption{Images are misclassified based on visual representation, and still misclassified based on both visual representation and motion representation.}
\label{viz2}
\end{figure}

\begin{figure}[t]
\centering
\includegraphics[width=0.9\columnwidth]{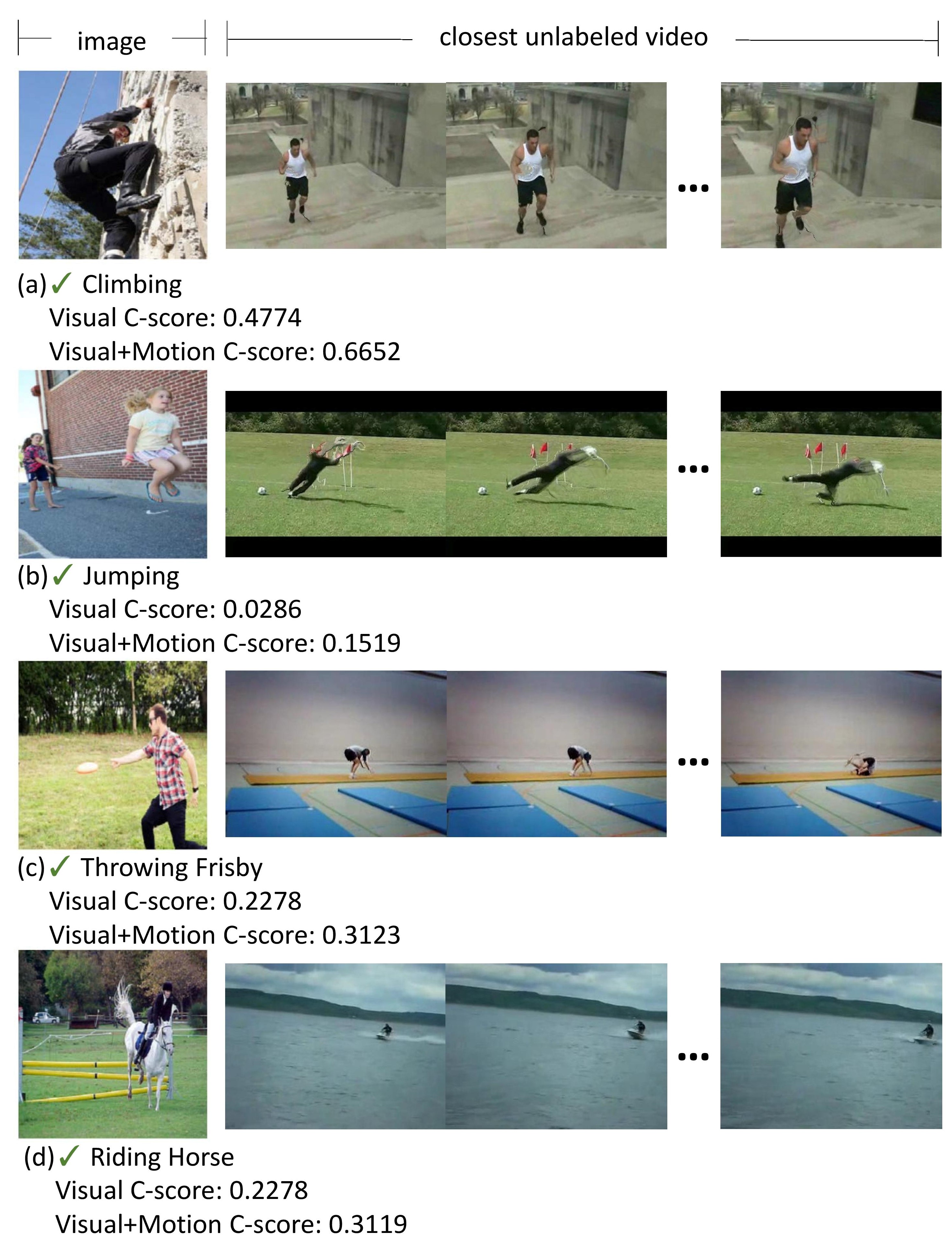}
\caption{Images are already correctly classified based on visual representation, but the confidence score corresponding to the correct category is improved based on both visual representation and motion representation. }
\label{viz3}
\end{figure}

\section{Qualitative Analyses of Motion Representation} 
In this section, we tend to elaborate the effectiveness of auxiliary motion representation in a qualitative way.
For compact description, we call Visual Representation (\emph{resp.}, Motion Representation)  VR (\emph{resp.}, MR) for short. We explore three types of cases: 1) images are misclassified based on VR, but are correctly classified based on both VR and MR; 2) images are misclassified based on VR, and still misclassified based on both VR and MR; 3) images are already correctly classified based on VR, but the confidence score corresponding to the correct category is improved based on both VR and MR. 

In the main paper, we have studied the first type of cases. Now, we study the second and the third types of cases are shown in Figure \ref{viz2} and \ref{viz3} respectively. In the second type of cases, as shown in Figure~\ref{viz2}, we find that auxiliary MR may be not helpful when the motion information in MR is not the emphasis of the image. For example, the image in (a) from the category ``applauding" is misclassified as ``brushing teeth" based on VR, because some grinning people show their teeth in the image. The motion information in MR is ``people laughing" without focusing on the hand movement, so it does not help correct the prediction. In the third type of cases, as shown in Figure~\ref{viz3}, we can see that auxiliary MR can help make more confident prediction. For example, the confidence score in (a) is raised from 0.4774 to 0.6652.

\begin{table}[t!]
\caption{Accuracies (\%) of different methods in two experimental settings. The best results are denoted in boldface.}
\centering
\small
\begin{tabular}{c|c c} 
\toprule[1pt]
 Datasets & UCF101\(\rightarrow\)VOC & HMDB\(\rightarrow\)VOC\\ \hline
UNREG  & 16.44 & 16.44 \\
O3N &16.63 & 17.67\\
Misra \emph{etal.}  & 17.40 & 20.11 \\
OPN & 18.63 & 19.00\\
Srivastava \emph{etal.} & 18.56 & 20.89\\
Purushwalkam \emph{etal.} & 16.79 & 16.85\\
SSFA  &19.78 &19.00\\
\hline
Carl \emph{etal.} &18.83 &18.72\\
Gao \emph{etal.} (2s)  &17.75 &17.94 \\
Gao \emph{etal.}  &17.62 &19.72 \\
Pintea \emph{etal.}&18.39 & 17.50\\
Jacob \emph{etal.} &17.94 & 19.11\\
\hline
\textbf{Ours} &\textbf{21.06} &\textbf{22.22}\\
\bottomrule[1pt]
\end{tabular}
\label{table:2}
\end{table}

\section{Impact of Different Network Structures}
To demonstrate the generalizability of our framework, we replace the AlexNet backbone with VGG16~\cite{simonyan2014very}. By taking two experimental settings (\emph{i.e.}, UCF101\(\rightarrow\)VOC and HMDB\(\rightarrow\)VOC) as examples, we re-evaluate all methods and report the results in Table \ref{table:2}, from which we have similar observations to AlexNet. Our framework still achieves the best results in two settings, which indicates that our framework can generalize to different network structures.

\bibliographystyle{aaai}
\small
\bibliography{aaai}

\begin{thebibliography}{}

\bibitem[\protect\citeauthoryear{Carreira and Zisserman}{2017}]{i3d}
Carreira, J., and Zisserman, A.
\newblock 2017.
\newblock Quo vadis, action recognition? {A} new model and the kinetics
  dataset.
\newblock In {\em CVPR}.

\bibitem[\protect\citeauthoryear{Chung \bgroup et al\mbox.\egroup }{2015}]{GRU}
Chung, J.; G{\"{u}}l{\c{c}}ehre, {\c{C}}.; Cho, K.; and Bengio, Y.
\newblock 2015.
\newblock Gated feedback recurrent neural networks.
\newblock In {\em ICML}.

\bibitem[\protect\citeauthoryear{Deng \bgroup et al\mbox.\egroup
  }{2009}]{deng2009imagenet}
Deng, J.; Dong, W.; Socher, R.; Li, L.-J.; Li, K.; and Fei-Fei, L.
\newblock 2009.
\newblock Imagenet: A large-scale hierarchical image database.
\newblock In {\em CVPR}.

\bibitem[\protect\citeauthoryear{Doersch, Gupta, and
  Efros}{2015}]{doersch2015unsupervised}
Doersch, C.; Gupta, A.; and Efros, A.~A.
\newblock 2015.
\newblock Unsupervised visual representation learning by context prediction.
\newblock In {\em ICCV}.

\bibitem[\protect\citeauthoryear{Everingham \bgroup et al\mbox.\egroup
  }{2010}]{VOC}
Everingham, M.; Gool, L. J.~V.; Williams, C. K.~I.; Winn, J.~M.; and Zisserman,
  A.
\newblock 2010.
\newblock The pascal visual object classes {(VOC)} challenge.
\newblock {\em IJCV} 88(2):303--338.

\bibitem[\protect\citeauthoryear{Fernando \bgroup et al\mbox.\egroup
  }{2015}]{Darwin}
Fernando, B.; Gavves, E.; M., J.~O.; Ghodrati, A.; and Tuytelaars, T.
\newblock 2015.
\newblock Modeling video evolution for action recognition.
\newblock In {\em CVPR}.

\bibitem[\protect\citeauthoryear{Fernando \bgroup et al\mbox.\egroup
  }{2017}]{odd-one-out}
Fernando, B.; Bilen, H.; Gavves, E.; and Gould, S.
\newblock 2017.
\newblock Self-supervised video representation learning with odd-one-out
  networks.
\newblock In {\em CVPR}.

\bibitem[\protect\citeauthoryear{Gao, Xiong, and Grauman}{2018}]{im2flow}
Gao, R.; Xiong, B.; and Grauman, K.
\newblock 2018.
\newblock Im2flow: Motion hallucination from static images for action
  recognition.
\newblock In {\em CVPR}.

\bibitem[\protect\citeauthoryear{Gidaris, Singh, and
  Komodakis}{2018}]{gidaris2018unsupervised}
Gidaris, S.; Singh, P.; and Komodakis, N.
\newblock 2018.
\newblock Unsupervised representation learning by predicting image rotations.
\newblock {\em ICLR}.

\bibitem[\protect\citeauthoryear{Hadsell, Chopra, and LeCun}{2006a}]{SFA2}
Hadsell, R.; Chopra, S.; and LeCun, Y.
\newblock 2006a.
\newblock Dimensionality reduction by learning an invariant mapping.
\newblock In {\em Computer Society Conference on Computer Vision and Pattern
  Recognition},  1735--1742.

\bibitem[\protect\citeauthoryear{Hadsell, Chopra, and LeCun}{2006b}]{contrast}
Hadsell, R.; Chopra, S.; and LeCun, Y.
\newblock 2006b.
\newblock Dimensionality reduction by learning an invariant mapping.
\newblock In {\em CVPR}.

\bibitem[\protect\citeauthoryear{Jain and Dubes}{1988}]{kmeans}
Jain, A.~K., and Dubes, R.~C.
\newblock 1988.
\newblock {\em Algorithms for Clustering Data}.
\newblock Prentice-Hall.

\bibitem[\protect\citeauthoryear{Jayaraman and Grauman}{2016}]{slow-steady}
Jayaraman, D., and Grauman, K.
\newblock 2016.
\newblock Slow and steady feature analysis: Higher order temporal coherence in
  video.
\newblock In {\em CVPR},  3852--3861.

\bibitem[\protect\citeauthoryear{J{\'{e}}gou \bgroup et al\mbox.\egroup
  }{2012}]{vlad}
J{\'{e}}gou, H.; Perronnin, F.; Douze, M.; S{\'{a}}nchez, J.; P{\'{e}}rez, P.;
  and Schmid, C.
\newblock 2012.
\newblock Aggregating local image descriptors into compact codes.
\newblock {\em T-PAMI} 34(9):1704--1716.

\bibitem[\protect\citeauthoryear{Krizhevsky, Sutskever, and
  Hinton}{2017}]{AlexNet}
Krizhevsky, A.; Sutskever, I.; and Hinton, G.~E.
\newblock 2017.
\newblock Imagenet classification with deep convolutional neural networks.
\newblock {\em Commun. {ACM}} 60(6):84--90.

\bibitem[\protect\citeauthoryear{Kuehne \bgroup et al\mbox.\egroup
  }{2011}]{HMDB}
Kuehne, H.; Jhuang, H.; Garrote, E.; Poggio, T.~A.; and Serre, T.
\newblock 2011.
\newblock {HMDB:} {A} large video database for human motion recognition.
\newblock In {\em ICCV}.

\bibitem[\protect\citeauthoryear{Lee \bgroup et al\mbox.\egroup
  }{2017}]{sort-seq}
Lee, H.; Huang, J.; Singh, M.; and Yang, M.
\newblock 2017.
\newblock Unsupervised representation learning by sorting sequences.
\newblock In {\em ICCV},  667--676.

\bibitem[\protect\citeauthoryear{Liu}{2009}]{course2fine}
Liu, C.
\newblock 2009.
\newblock {\em Exploring new representations and applications for motion
  analysis}.
\newblock Ph.D. Dissertation, Massachusetts Institute of Technology, Cambridge,
  MA, {USA}.

\bibitem[\protect\citeauthoryear{Mezghiche, Melkemi, and Foufou}{2014}]{SC}
Mezghiche, K.~M.; Melkemi, K.~E.; and Foufou, S.
\newblock 2014.
\newblock Matching with quantum genetic algorithm and shape contexts.
\newblock In {\em AICCSA}.

\bibitem[\protect\citeauthoryear{Misra, Zitnick, and
  Hebert}{2016}]{shuffle-learn}
Misra, I.; Zitnick, C.~L.; and Hebert, M.
\newblock 2016.
\newblock Shuffle and learn: Unsupervised learning using temporal order
  verification.
\newblock In {\em ECCV}.

\bibitem[\protect\citeauthoryear{Mobahi, Collobert, and Weston}{2009}]{SFA1}
Mobahi, H.; Collobert, R.; and Weston, J.
\newblock 2009.
\newblock Deep learning from temporal coherence in video.
\newblock In {\em ICML}.

\bibitem[\protect\citeauthoryear{Noroozi and
  Favaro}{2016}]{noroozi2016unsupervised}
Noroozi, M., and Favaro, P.
\newblock 2016.
\newblock Unsupervised learning of visual representations by solving jigsaw
  puzzles.
\newblock In {\em ECCV}.

\bibitem[\protect\citeauthoryear{Pintea, van Gemert, and Smeulders}{2014}]{rf}
Pintea, S.~L.; van Gemert, J.~C.; and Smeulders, A. W.~M.
\newblock 2014.
\newblock D{\'{e}}j{\`{a}} vu: - motion prediction in static images.
\newblock In {\em ECCV},  172--187.

\bibitem[\protect\citeauthoryear{Purushwalkam and Gupta}{2016}]{pose-action}
Purushwalkam, S., and Gupta, A.
\newblock 2016.
\newblock Pose from action: Unsupervised learning of pose features based on
  motion.
\newblock {\em CoRR} abs/1609.05420.

\bibitem[\protect\citeauthoryear{Schmid \bgroup et al\mbox.\egroup
  }{2013}]{bio2}
Schmid, M.~C.; Schmiedt, J.~T.; Peters, A.~J.; Saunders, R.~C.; Maier, A.; and
  Leopold, D.~A.
\newblock 2013.
\newblock Motion-sensitive responses in visual area v4 in the absence of
  primary visual cortex.
\newblock {\em The Journal of neuroscience : the official journal of the
  Society for Neuroscience} 33(48):18740--5.

\bibitem[\protect\citeauthoryear{Sharma, Jurie, and Schmid}{2012}]{human-body}
Sharma, G.; Jurie, F.; and Schmid, C.
\newblock 2012.
\newblock Discriminative spatial saliency for image classification.
\newblock In {\em CVPR}.

\bibitem[\protect\citeauthoryear{Shima \bgroup et al\mbox.\egroup
  }{1986}]{otsu}
Shima, Y.; Kashioka, S.; Kato, K.; and Ejiri, M.
\newblock 1986.
\newblock An automatic visual inspection method for a plastic surface based on
  image partitioning and gray-level histograms.
\newblock {\em Systems and Computers in Japan} 17(5):54--63.

\bibitem[\protect\citeauthoryear{Simonyan and Zisserman}{2014a}]{two-stream}
Simonyan, K., and Zisserman, A.
\newblock 2014a.
\newblock Two-stream convolutional networks for action recognition in videos.
\newblock In {\em NIPS}.

\bibitem[\protect\citeauthoryear{Simonyan and
  Zisserman}{2014b}]{simonyan2014very}
Simonyan, K., and Zisserman, A.
\newblock 2014b.
\newblock Very deep convolutional networks for large-scale image recognition.
\newblock {\em arXiv preprint arXiv:1409.1556}.

\bibitem[\protect\citeauthoryear{Soomro, Zamir, and Shah}{2012}]{UCF101}
Soomro, K.; Zamir, A.~R.; and Shah, M.
\newblock 2012.
\newblock {UCF101:} {A} dataset of 101 human actions classes from videos in the
  wild.
\newblock {\em CoRR} abs/1212.0402.

\bibitem[\protect\citeauthoryear{Srivastava, Mansimov, and
  Salakhutdinov}{2015}]{lstm}
Srivastava, N.; Mansimov, E.; and Salakhutdinov, R.
\newblock 2015.
\newblock Unsupervised learning of video representations using lstms.
\newblock In {\em ICML}.

\bibitem[\protect\citeauthoryear{Tran \bgroup et al\mbox.\egroup
  }{2015}]{deepnet2}
Tran, D.; Bourdev, L.~D.; Fergus, R.; Torresani, L.; and Paluri, M.
\newblock 2015.
\newblock Learning spatiotemporal features with 3d convolutional networks.
\newblock In {\em ICCV}.

\bibitem[\protect\citeauthoryear{Vondrick, Pirsiavash, and
  Torralba}{2016}]{reg}
Vondrick, C.; Pirsiavash, H.; and Torralba, A.
\newblock 2016.
\newblock Anticipating visual representations from unlabeled video.
\newblock In {\em CVPR}.

\bibitem[\protect\citeauthoryear{Walker, Gupta, and
  Hebert}{2015}]{dense-optical}
Walker, J.; Gupta, A.; and Hebert, M.
\newblock 2015.
\newblock Dense optical flow prediction from a static image.
\newblock In {\em ICCV}.

\bibitem[\protect\citeauthoryear{Wang and Gupta}{2015}]{Wang}
Wang, X., and Gupta, A.
\newblock 2015.
\newblock Unsupervised learning of visual representations using videos.
\newblock In {\em ICCV},  2794--2802.

\bibitem[\protect\citeauthoryear{Wiskott and Sejnowski}{2002}]{SFA}
Wiskott, L., and Sejnowski, T.~J.
\newblock 2002.
\newblock Slow feature analysis: Unsupervised learning of invariances.
\newblock {\em Neural Computation} 14(4):715--770.

\bibitem[\protect\citeauthoryear{Yao \bgroup et al\mbox.\egroup }{2011}]{S40}
Yao, B.; Jiang, X.; Khosla, A.; Lin, A.~L.; Guibas, L.~J.; and Li, F.
\newblock 2011.
\newblock Human action recognition by learning bases of action attributes and
  parts.
\newblock In {\em ICCV}.

\bibitem[\protect\citeauthoryear{Zhang \bgroup et al\mbox.\egroup
  }{2016}]{zhang2016action}
Zhang, Y.; Cheng, L.; Wu, J.; Cai, J.; Do, M.~N.; and Lu, J.
\newblock 2016.
\newblock Action recognition in still images with minimum annotation efforts.
\newblock {\em T-IP} 25(11):5479--5490.

\bibitem[\protect\citeauthoryear{Zhang, Isola, and
  Efros}{2017}]{zhang2017split}
Zhang, R.; Isola, P.; and Efros, A.~A.
\newblock 2017.
\newblock Split-brain autoencoders: Unsupervised learning by cross-channel
  prediction.
\newblock In {\em CVPR}.

\bibitem[\protect\citeauthoryear{Zheng \bgroup et al\mbox.\egroup
  }{2012}]{context-scence}
Zheng, Y.; Zhang, Y.; Li, X.; and Liu, B.
\newblock 2012.
\newblock Action recognition in still images using a combination of human pose
  and context information.
\newblock In {\em ICIP}.

\end{thebibliography}
\end{document}